\documentclass[twoside,11pt]{article}
\usepackage{pgm}
\usepackage{enumitem,kantlipsum, float}

\usepackage[utf8]{inputenc}
\inputencoding{utf8}

\usepackage{hyperref,color,soul,booktabs}
\setulcolor{blue}
\newcommand{\BibTeX}{\textsc{B\kern-0.1emi\kern-0.017emb}\kern-0.15em\TeX}
\usepackage{amsmath, amssymb, txfonts}

\newcommand{\calT}{\mathcal{T}}
\newcommand{\calC}{\mathcal{C}}
\newcommand{\calS}{\mathcal{S}}
\newcommand{\calG}{\mathcal{G}}
\newcommand{\mtrian}{\mathrel{\raisebox{-0.1ex}{%
\scalebox{0.8}[0.6]{$\vartriangle$}}}}
\newcommand{\defineas}{\overset{\mtrian}{=}} 
\usepackage{verbatim}

\ShortHeadings{Staged Tree Algorithm}{Shenvi and Smith}

\begin{document}

\title{Constructing a Chain Event Graph from a Staged Tree}

\author{\Name{Aditi Shenvi $^{1,2}$} \Email{a.shenvi@warwick.ac.uk}\and
   \Name{Jim Q. Smith $^{1,2}$} \Email{j.q.smith@warwick.ac.uk}\\
   \addr $^{1\,}$ University of Warwick, Coventry, United Kingdom \\
    $^{2\,}$ The Alan Turing Institute, London, United Kingdom}

\maketitle

\begin{abstract}
Chain Event Graphs (CEGs) are a recent family of probabilistic graphical models - a generalisation of Bayesian Networks - providing an explicit representation of structural zeros, structural missing values and context-specific conditional independences within their graph topology. A CEG is constructed from an event tree through a sequence of transformations beginning with the colouring of the vertices of the event tree to identify one-step transition symmetries. This coloured event tree, also known as a staged tree, is the output of the learning algorithms used for this family. Surprisingly, no general algorithm has yet been devised that automatically transforms any staged tree into a CEG representation. In this paper we provide a simple iterative backward algorithm for this transformation. Additionally, we show that no information is lost from transforming a staged tree into a CEG. Finally, we demonstrate that with an optimal stopping criterion, our algorithm is more efficient than the generalisation of a special case presented in \cite{silander2013dynamic}. We also provide Python code using this algorithm to obtain a CEG from any staged tree along with the functionality to add edges with sampling zeros.  

\end{abstract}

\section{Introduction}
Many real-world processes contain non-symmetric sample space structures. Examples of such processes can be frequently found in public health, medicine, risk analysis and policing (see \cite{collazo2018chain}). Such asymmetries may arise due to the existence of structural zeros and structural missing values (collectively referred here as structural asymmetries) in the sample space of a variable conditional on the realisation of other variable(s). A structural zero refers to observing zero frequencies for a count variable or a category of a categorical variable when a non-zero observation is a logical impossibility rather than a sampling limitation (e.g. days or amount as low, medium, high of alcohol consumption by teetotallers). Structural missing value are observations which are missing as they are not defined for a subset of the individuals/units (e.g. variables relating to post-operative health of individuals who had the illness but weren't operated). It is easy to see how such asymmetries may give rise to context-specific conditional independences which are independence relationships of the form $X \Perp Y|Z = z_1$ but $X \not \Perp Y | Z = z_2$ where $\Perp$ stands for probabilistic independence and the vertical bar shows conditioning variables on the right. In fact, context-specific independences regularly arise naturally in many applications \citep{zhang1999role}.

Graphical models such as Bayesian Networks (BNs) are unable to fully describe asymmetric processes. They are primarily stymied in this respect as they force the process description on a set of variables that are defined \textit{a priori}. Indeed, in order to scale up BN methodologies to large problems, good BN software contain functions that copy parts of one conditional probability table to another. Thus BNs implicitly embed context-specific independences through probability assignments within their conditional probability tables. However, this structural information is never explicitly represented in their topologies. Uncovering these independences requires serious modifications (typically involving trees in some form) to their standard representation and/or inferential process \citep{boutilier1996context, zhang1999role, jabbari2018instance}. Additionally, structural zeros too are hidden away in their conditional probability tables. 

Chain Event Graphs (CEGs) are a family of probabilistic graphical models whose graphical representation make structural asymmetries and context-specific conditional independences explicit \citep{collazo2018chain}. CEGs contain the class of finite discrete BNs as a special case \citep{smith2008conditional}. They are constructed from \textit{event trees} which provide a natural and intuitive framework for describing the unfolding of a process through a sequence of \textit{events}. Although the size of an event tree increases linearly with the number of events involved in the evolution of the process which may become unwieldy for large complex processes, they are nonetheless easy for the statistician to transparently elicit from the natural language descriptions of a domain expert. Embedding structural asymmetries within an event tree is a matter of simply not drawing the corresponding branch in the tree \citep{shenvi2018modelling}. However, a more compact representation of an event tree while retaining its properties and transparency is desirable. A CEG provides such a compact representation. Hence it is a powerful modelling tool for processes exhibiting significant asymmetries, particularly in domains such as medicine \citep{barclay2013refining}, public health \citep{shenvi2018modelling}, forensic science \citep{collazo2018chain} where experts often offer event based descriptions of processes. 

To obtain a CEG, an event tree is first transformed into a \textit{staged tree} by colouring its vertices to represent symmetries within its structure. The vertices of the staged tree are then merged to provide a more concise representation of these symmetries in the form of the graph of a CEG. Such a transformation results in a much simpler graph often with an order of magnitude fewer vertices and edges than the generating tree. Like an event tree, a CEG also describes a process through a sequence of events and thus inherits the ability to graphically represent structural asymmetries. A CEG representation is especially useful because various implicit conditional independences, including of the context-specific nature, hidden within the patterns of colouring of the tree can be read directly from its topology using sets of events called \textit{cuts} and \textit{fine cuts} \citep{smith2008conditional}. 

Several fast learning algorithms now exist for the CEG \citep{freeman2011bayesian, silander2013dynamic, cowell2014causal}. The output of these algorithms is a staged tree. A staged tree typically must go through a sequence of non-trivial transformations before it represents the graph of a CEG. In fact, a CEG is uniquely defined by its staged tree, and we show that the staged tree can be recovered from the graph of the CEG alone.

\cite{silander2013dynamic} present an algorithm to transform a \textit{stratified} staged tree into a \textit{stratified} CEG (SCEG). A stratified staged tree/ SCEG is one in which events broadly corresponding to the same variable are at the same distance from a leaf/ the sink. Intuitively this corresponds to there being no events which become redundant conditional on the past events that have occurred. SCEGs have been studied extensively as any process that can be represented by a finite discrete BN can also be represented within this wider class. In particular, the advantages of the CEG over a BN can be demonstrated \citep{barclay2013refining}. However, we are increasingly finding many applications where the CEG representation is not stratified \citep{shenvi2018modelling, shenvi2019bayesian}. So it is timely that automatic algorithms are available to make this transformation for \textit{any} staged tree.

The contribution of our paper is threefold. First we provide an algorithm that can transform any staged tree into a CEG and provide an optimal stopping criterion for this algorithm. Secondly, we prove that the transformation of a staged tree into a CEG does not lead to the loss of any information. Lastly, we provide Python code (\url{https://github.com/ashenvi10/Chain-Event-Graphs}) that obtains a staged tree using an Agglomerative Hierarchical Clustering (AHC) algorithm and then transforms it into a CEG using our algorithm. Unlike the existing `ceg' \citep{ceg_package} and `stagedtrees' \citep{carli2020r} \textsf{R} packages, our code is not restricted to SCEGs and it also allows manual addition of edges with sampling zeros.

In Section \ref{notation} we review the notation and the preliminary concepts. In Section \ref{algorithm} we present a simple recursive backward algorithm - coded within supporting software - that can construct a CEG from any staged tree. Here we also prove some properties of the algorithm and of the transformation itself. In Section \ref{experiments} we compare an adapted version of the algorithm presented in \cite{silander2013dynamic} to our algorithm. We conclude the paper with a short discussion in Section \ref{discussion}.

\section{Notation and Preliminaries} \label{notation}
\begin{figure}[ht]
    \centering
    \includegraphics[trim = 9cm 1.5cm 3cm 3cm, scale = 0.48 ]{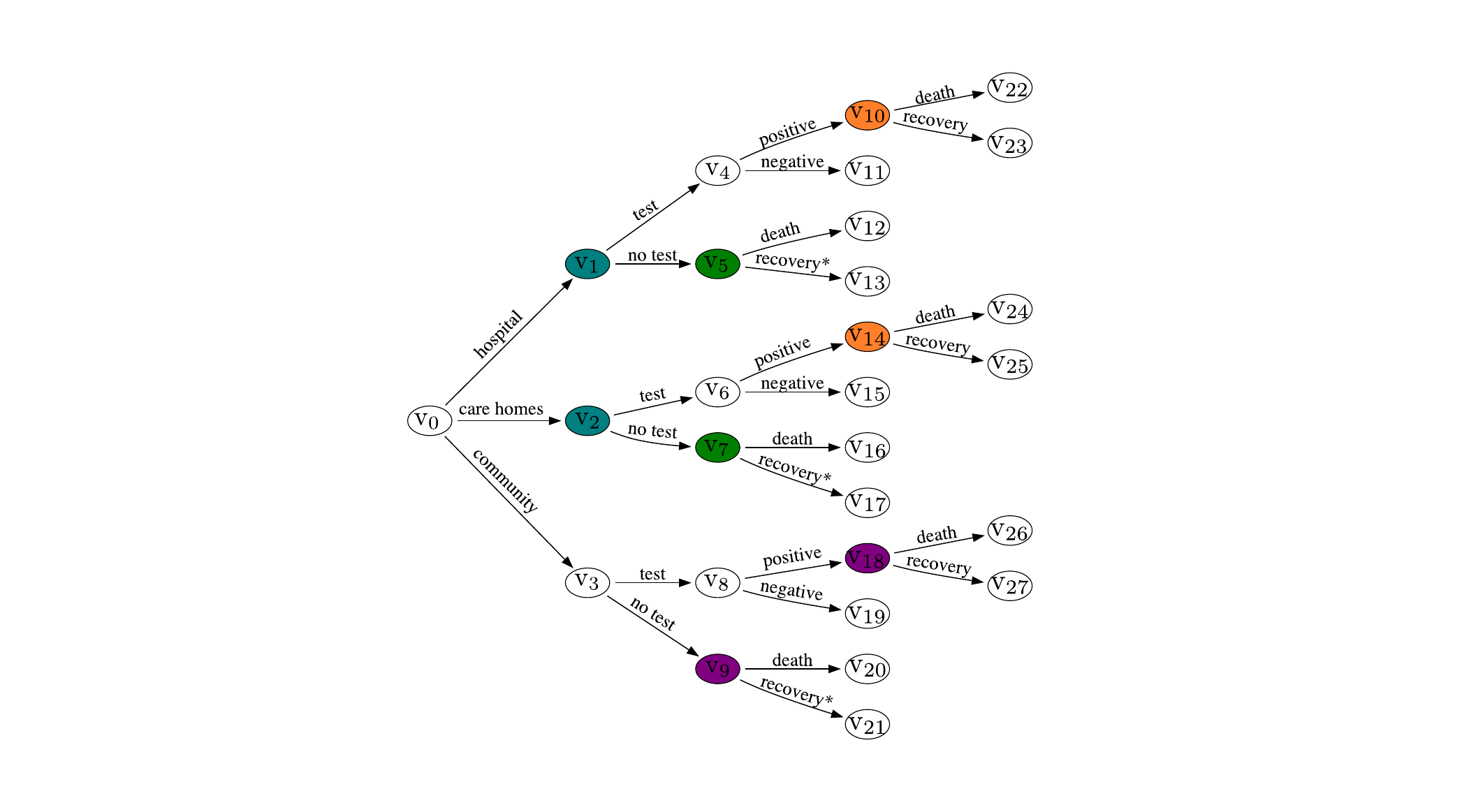}
    \caption{Staged tree for Example \ref{example_topical}}
    \label{fig:topical_staged_tree}
\end{figure}

\begin{figure}[ht]
    \centering
    \includegraphics[trim = 7cm 1.2cm 3cm 2cm, scale = 0.29]{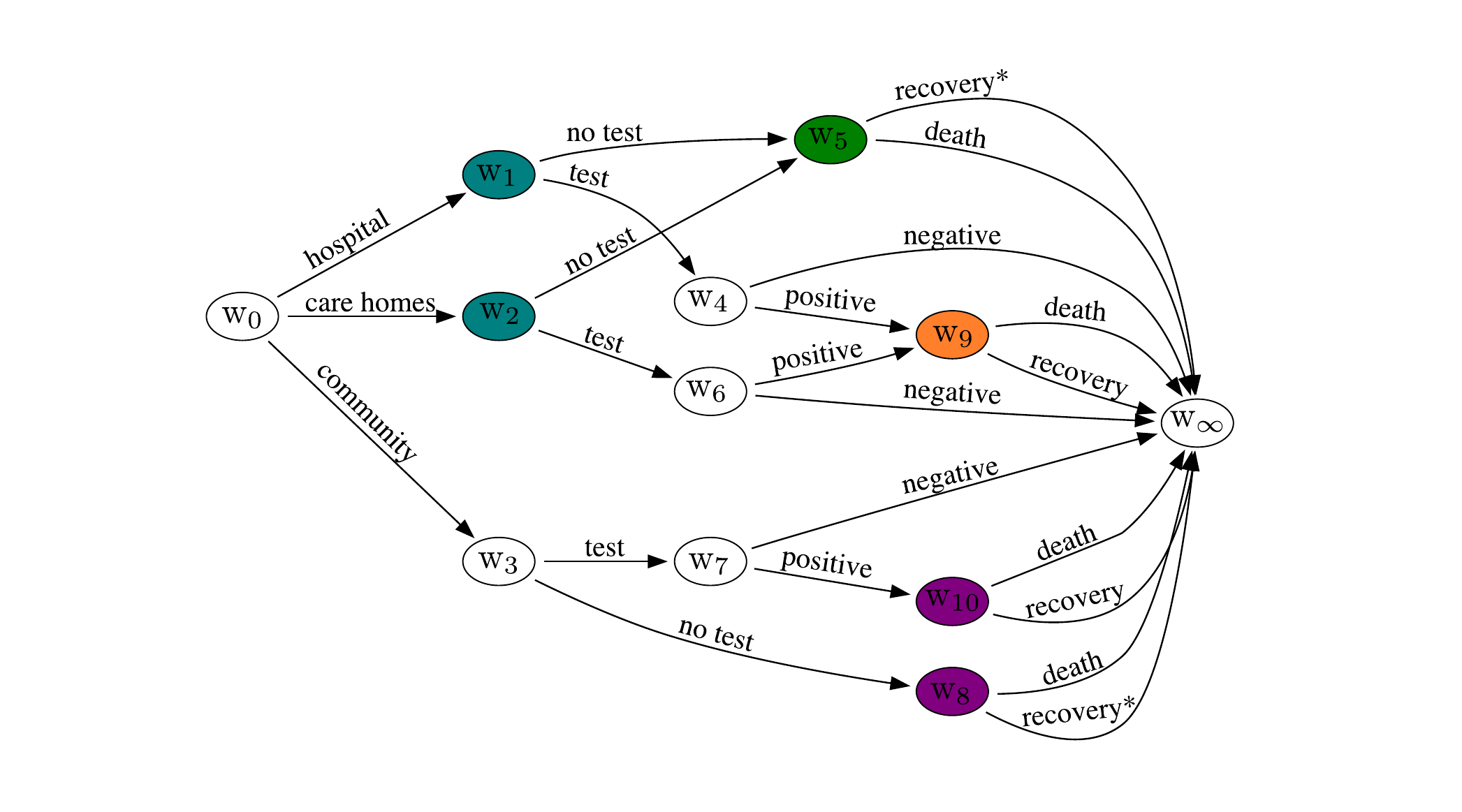}
    \caption{CEG for Example \ref{example_topical}}
    \label{fig:topical_ceg}
\end{figure}

A CEG construction begins by eliciting an event tree of a process either from a domain expert or from existing literature. Alternatively, it can be constructed directly from data. Below we outline the transformations an event tree goes through to obtain the graph of a CEG:
\begin{itemize}
    \itemsep0em
    \item Vertices in the event tree whose one step ahead evolutions, i.e. conditional transition probabilities, are equivalent are assigned the same colour to indicate this symmetry;
    \item Vertices whose rooted subtrees (the subtree formed by considering that vertex as the root) are isomorphic - in the structure and colour preserving sense - are merged into a single vertex which retains the colouring of its merged vertices;
    \item All the leaves of the tree are merged into a single vertex called the sink.
\end{itemize}
\begin{example} \label{example_topical}
Here we consider a simplified topical example. The staged tree in Figure \ref{fig:topical_staged_tree} shows a hypothesised example of testing for a certain disease available to individuals exhibiting symptoms in three different settings: hospitals, care homes and in the general community. For simplicity, we assume here that the test is 100\% sensitive and specific, and that we are only interested in the outcomes related to the disease. By ``recovery$^*$" we collectively refer to those who recover and those who never had the disease. We further assume that death can only be caused by the disease in the time period considered. The coloured vertices represent equivalence of their conditional transition probabilities. For instance, the probability of dying is the same for individuals in hospitals and care homes who exhibit symptoms but do not get a test. The CEG for this staged tree is shown in Figure \ref{fig:topical_ceg}. It is not hard to see how this tree can be refined to be more realistic.
\end{example}
Let $\calT$ denote an event tree with a finite vertex set $V(\calT)$ and an edge set $E(\calT)$. An edge $e \in E(\calT)$ from vertex $v$ to vertex $v'$ with edge label $l$ is an ordered triple given by $(v, v', l)$. Denote by $L(\calT)$ the set of leaves in $\calT$. The non-leaf vertices in $\calT$ are called \textit{situations} and their set is denoted by $S(\calT) = V(\calT) \backslash L(\calT)$. The set of children of a vertex $v$ are denoted by $\textmd{ch}(v)$. Let $\Phi_\calT = \{\pmb{\theta}_v |v \in V(\calT)\}$ where $\pmb{\theta_v} = (\theta(e) | e = (v, v', l) \in E(\calT), v' \in V(\calT))$ denotes the parameters for each vertex $v \in V(\calT)$.

Two situations $v$ and $v'$ are said to be in the same \textit{stage} whenever $\pmb{\theta}_v = \pmb{\theta}_{v'}$ and if $\theta(e) = \theta(e')$ then $e = (v, \cdot, l)$ and $e' = (v', \cdot, l)$ for edge $e$ emanating from $v$ and $e'$ emanating from $v'$. The latter condition states that the edges emanating from situations in the same stage which have the same estimated conditional transition probability must also share the same edge label. Note that when edge labels are not fixed, this condition is relaxed. In this case, edges of vertices in the same stage are coloured to represent which edges share the same conditional transition probabilities. This allows the statistician and domain expert to retrospectively assign labels to events which have the same meaning but which could have initially been assigned different labels.
\setcounter{example}{0}
\begin{example}(Continued)
The domain expert may decide that the edge labels ``recovery" and ``recovery$^*$" can be treated as equivalent. Then situations $v_9$ and $v_{18}$ would be in the same position.
\end{example}

The collection of stages $\mathbb{U}$ partitions $V(\calT)$. Each stage $u \in \mathbb{U}$ is a set of situations in $V(\calT)$ that belong to the stage $u$. Stage memberships are represented by colouring the situations of $\calT$ such that each stage $u \in \mathbb{U}$ is represented by a unique colour. An event tree whose situations are coloured according to their stage memberships is called a \textit{staged tree} and is denoted by $\calS$ \footnote{Note that for simplicity, like in Figures \ref{fig:topical_staged_tree} and \ref{fig:topical_ceg}, the colouring of the trivial stages may be suppressed.}. Situations in the staged tree whose rooted subtrees are isomorphic \footnote{In this paper isomorphism is in a structure and colouring preserving sense.} have equivalent sets of parameters. That is, for two isomorphic subtrees $\calS'$ and $\calS''$ rooted at $v$ and $v'$, $\Theta_{\calS'} = \Theta_{\calS''}$. In a non-technical sense, this implies that $v$ and $v'$ have identical future evolutions. Situations whose rooted subtrees are isomorphic belong to the same \textit{position}. The collection of positions $\mathbb{W}$ is a finer partition of $V(\calT)$ and each position $w \in \mathbb{W}$ is a set of situations of $V(\calT)$ that belong to the position $w$. Merging the situations in $\calS$ which are in the same position and collecting all the leaves in $L(\calS)$ into a sink node denoted by $w_\infty$ result in the graph of a CEG $\calC$ for the process being modelled. Thus a CEG is uniquely defined by its staged tree, or in other words, it is uniquely defined by the pair $(\calT, \mathbb{U})$ where $\calT$ is its underlying event tree and $\mathbb{U}$ is the set of stages. 
\begin{definition}[Chain Event Graph]
A Chain Event Graph (CEG) $\calC = (V(\calC), E(\calC))$ of a process represented by a staged tree $\calS = (V(\calS), E(\calS))$ with set of parameters $\Theta_\calS$ is a directed acyclic graph with $V(\calC) = W \cup w_\infty$ where $W$ is a set constructed by choosing a representative situation from each set in the collection \footnote{Notice that $\mathbb{U}$ and $\mathbb{W}$ are sets of sets and to disambiguate, we refer to sets of sets as collections in this paper.} $\mathbb{W}$. The edges in $\calC$ are constructed as follows: For a $w \in W$, create an edge $(w, w', l) \in E(\calC)$ for every edge $(w, s', l) \in E(\calS)$,  with $\pmb{\theta}_w^{\calC} = \pmb{\theta}_w^{\calS}$ where $s'$ belongs to a set in $\mathbb{W}$ which is represented by $w'$ in $W$. Additionally, $w \in V(\calC)$ retains the colouring of $w \in V(\calS)$.
\end{definition}

A \textit{floret} of a vertex $v$ in a directed graph is denoted by $F(v) = (V(F(v), E(F(v)))$ where $V(F(v)) = \{v \cup \textmd{ch}(v)\}$ and $E(F(v))$ is the set of edges induced by $V(F(v))$ in the graph. Denote the set of root-to-sink (root-to-leaf) paths in a CEG $\calC$ (event tree $\calT$ /staged tree $\calS$) by $\calC_\Lambda$ ($\calT_\Lambda$ / $\calS_\Lambda$) where a path is a sequence of tuples of the form \textit{(`vertex colour', `edge label')} from the root vertex to the sink following the directed edges. Say that an event tree, staged tree or a CEG is \textit{stratified} whenever the vertices representing the same type of event (e.g. severity of illness) have the same number of edges between them and the root vertex along any path connecting them, and otherwise say it is \textit{non-stratified}. Non-stratified CEGs provide a more realistic representation of a wide range of processes containing structural asymmetries  (see e.g. \cite{shenvi2018modelling, shenvi2019bayesian}).

\subsection{Why not just Staged Trees?} \label{staged_trees}

Staged trees are a graphical representation of a parametric statistical model and encapsulate within their colouring conditional independence information about the events describing a process \citep{gorgen2016differential, gorgen2018equivalence}. So why do we need CEGs when staged trees are themselves powerful tools? 

While we show that staged tree and CEG representations are equivalent, the graph of a CEG is simpler and more compact. Typically, a CEG contains far fewer vertices and edges than its corresponding staged tree. Let  $V_{k} \subseteq V(\calT)$ denote the vertices of an event tree $\calT$ with $k$ outgoing edges and let $n_{k}=\abs{V_{k}}$. Then $\mathcal{T}$ has $n(\mathcal{T})=\sum_{k=0}^{d}n_{k}$ vertices and $k(\mathcal{T})=\sum_{k=1}^{d}kn_{k}$ edges where $d = \textmd{max} \{k: n_k \geq 0\}$. When a CEG $\mathcal{C}$ partitions $V_{k}$ into $1\leq m_{k}\leq n_{k}$ positions, it is trivial to check that it has $n(\mathcal{C})=\sum_{k=1}^{d}m_{k}+1$ vertices (including the sink) and $k(\mathcal{C})=\sum_{k=1}^{d}km_{k}$ edges. So we have
\begin{align*}
n-1  & \leq n(\mathcal{T})-n(\mathcal{C})= \textstyle \sum_{k=1}^{d}\left(  n_{k}-m_{k}\right)
+n-1\leq \textstyle\sum_{k=0}^{d}\left(  n_{k}-1\right),  \\
0  & \leq k(\mathcal{T})-k(\mathcal{C})=\textstyle \sum_{k=1}^{d}k\left(  n_{k}-m_{k}\right)  \leq \textstyle \sum
_{k=1}^{d}k\left(  n_{k}-1\right),
\end{align*}

\noindent where $n = \abs{L(\calT)}$. Let $m = \textmd{max } \{\abs{\lambda} \,:\, \lambda \in \calT_\Lambda\}$ be the length (i.e. number of tuples) of the longest root-to-leaf path of $\mathcal{T}$. It is easy to check that $n(\mathcal{T})$ and $k(\mathcal{T})$ typically increase as a power of $m$, whilst when its CEG expresses many symmetries $n(\mathcal{C})$ and $k(\mathcal{C})$ increase linearly in $m$. In fact, for dynamic processes, the staged tree is infinite but the corresponding CEG might be finite \citep{shenvi2019bayesian}. Crucially, while there now exists a d-separation theorem (to be reported soon) for CEGs, such methodologies are yet to be developed for staged trees. Note that there is an interesting framework called conditional independence trees \citep{su2005representing, zhang2004conditional} which decompose decision trees into subtrees by exploiting the conditional independence relationships (including those of the context-specific nature) exhibited by the process. However, these are not yet fully developed and have been primarily used for improving prediction on classification problems.

\section{A Recursive Algorithm to Construct a CEG} \label{algorithm}
We present a simple recursive backward algorithm for constructing the graph of a CEG $\calC$ from any staged tree $\calS$ irrespective of whether it is stratified. While a variety of model selection techniques exist for the CEG family \citep{freeman2011bayesian, silander2013dynamic, cowell2014causal}, we do not discuss these in this paper. The outcome of any model selection algorithm for a CEG is a collection of stages $\mathbb{U}$ for its underlying event tree. The vertices of the event tree can be coloured according to $\mathbb{U}$; giving us the associated staged tree. Here we assume that we are only given the staged tree - obtained either as an output of a model selection algorithm or elicited by domain experts - from which we can deduce the collection of stages $\mathbb{U}$. The collection $\mathbb{U}$ and the topology of the staged tree are then used to iteratively identify the collection of positions. 

The recursion progressively melds situations together according to the position structure incrementally more distant from the leaves of the staged tree $\calS$. This produces a sequence of coloured graphs $\calG_{0} = \calS, \calG_{1},\ldots, \calG_{m} = \calC$ where $m$ is the depth of $\calS$. Each graph in the sequence has the same root-to-leaf/sink paths, that is $\calG_{0\Lambda} = \calG_{1\Lambda} = \ldots = \calG_{m\Lambda}$, and the following relationship holds
\begin{equation*}
    \abs{V(G_i)} \geq \abs{V(G_{i+1})}, \quad \abs{E(G_i)} \geq \abs{E(G_{i+1})}; \quad \quad i = 0, 1, \ldots, m-1.
\end{equation*}
We specify our construction by writing the vertex and edge sets of each graph $\calG_i$ as a function of the vertex and edge sets of the graph $\calG_{i-1}$. Note that the vertices in $\calG_i$ retain their colouring from the graph $\calG_{i-1}$. Henceforth, we will say $\calG_i = \calG_j$, $i \neq j$, whenever the two graphs $\calG_i$ and $\calG_j$ are isomorphic. Say that a vertex $v$ is at a distance $k$ from the sink vertex $w_\infty$ (or equivalently, a leaf in a tree) if the shortest directed path from $v$ to the sink (or a leaf) contains $k$ tuples. Let $V^{-k}$ be the set of vertices in a given graph such that every $v \in V^{-k}$ is at a distance of $k$ from the sink vertex $w_\infty$ (or a leaf) of the graph. We describe our iterative algorithm below.

\noindent \textbf{Step 1: Initialisation.} From $\calG_0 = \calS$ where $\calS$ is the staged tree, define the following: 
\begin{align*}
    \nu_1^- &\defineas L(\calG_0),\quad \quad \nu_1^+ \defineas \{w_\infty\},\\
    \epsilon_1^- &\defineas \{e \in E(\calG_0) \,:\, e = (v, v', l) \textmd{ where } v \in S(\calG_0), v' \in L(\calG_0)\}, \\
    \epsilon_1^+ &\defineas \{ \sigma_1(e) \,:\, e \in \epsilon_1^-\}, 
\end{align*}
\noindent where $\sigma_1(e) = \sigma_1(v, v', l) \defineas (v, w_\infty, l)$. Graph $\calG_1 = (V(\calG_1), E(\calG_1))$ where
\begin{align*}
    V(\calG_1) \defineas V(\calG_0) \backslash \nu_1^- \cup \nu_1^+, \quad \quad
    E(\calG_1) \defineas E(\calG_0) \backslash \epsilon_1^- \cup \epsilon_1^+.
\end{align*}

\noindent \textbf{Step 2: Generalisation.} To construct graph $\calG_{i}$ from $\calG_{i-1}$, $i \leq m$, proceed as follows:
\begin{enumerate}[leftmargin=*]
\itemsep0em
    \item Create a sub-collection $U_i = \{u_{1i},\ldots,u_{m_ii}\}$ informed by the collection of stages $\mathbb{U}$ such that each situation $v \in V^{-(i-1)}$ belongs to only one set $u_{ji} \in U_i$ for some $j = 1, \ldots, m_i$, and two situations $v, v' \in V^{-(i-1)}$ belong to the same set $u_{ji}$ if and only if there exists a stage $u \in \mathbb{U}$ such that $v, v' \in u$. Thus, the collection $U_i$ gives us the stage structure for the vertices in $V^{-(i-1)}$.
    
    \item Construct a collection $U_i^*$ such that each $u_{ji} \in U_i$ is replaced in $U_i^*$ by the sets $u_{ji}^1,\ldots,u_{ji}^{n_{ji}}$, $n_{ji} \geq 1$. Each situation $v \in u_{ji}$ belongs to only one set $u_{ji}^k$ for some $k = 1, \ldots, n_{ji}$, and two situations $v, v' \in u_{ji}$ belong to the same set $u_{ji}^k$ if and only if there exists an edge $(v', v'', l) \in E(\calG_{i-1})$ for every edge $(v, v'', l) \in E(\calG_{i-1})$. Thus, we have that $u_{ji}^k \cap u_{ji}^l = \emptyset$, $k \neq l$, $\cup_{k} u_{ji}^k = u_{ji}$, and $U_i^* =  \cup_j \cup_k u_{ji}^k$. The collection $U_i^*$ partitions the situations in $V^{-(i-1)}$ into positions. 
    
    \item Define the following terms for each $u_{ji}^k$, $j = 1,\ldots,m_i$, $k = 1,\ldots,n_{ji}$,
    \begin{align*}
        \nu^-(u_{ji}^k) &\defineas u_{ji}^k, \quad \quad  \nu^+(u_{ji}^k) \defineas \{v\} \textmd{ for some } v \in \nu^-(u_{ji}^k).
    \end{align*}
    We now define the following terms to enable us to construct the vertex and edge sets of $\calG_i$,
    \begin{align*}
        \nu_i^- &\defineas \cup_j \cup_k \nu^-(u_{ji}^k), \quad \quad  \nu^+_i \defineas \cup_j \cup_k \nu^+(u_{ji}^k),\\
        \epsilon_i^f &\defineas \{e \in E(\calG_{i-1}) \,:\, e = (v, v', l) \textmd{ where } v \in \nu_i^- \backslash \nu^+_i\}, \\
        \epsilon_i^b &\defineas \{e \in E(\calG_{i-1}) \,:\, e = (v, v', l) \textmd{ where } v' \in \nu_i^- \backslash \nu^+_i\}, \\
        \epsilon_i^- &\defineas \epsilon_i^f \cup \epsilon_i^b, \quad \quad \epsilon_i^+ \defineas \{\sigma_i(e) \,:\, e \in \epsilon_i^b\},
    \end{align*}
    \noindent where $\sigma_i(e) = \sigma_i(v, v', l) \defineas (v, v'', l)$ in which $v'' \in \nu^+(u_{ji}^k)$ for $v' \in \nu^-(u_{ji}^k)$, $k = 1, \ldots, n_{ji}$. Setting $V(\calG_i) \defineas V(\calG_{i-1}) \backslash \nu_i^- \cup \nu_i^+$ and
        $E(\calG_i) \defineas E(\calG_{i-1}) \backslash \epsilon_i^- \cup \epsilon_i^+$ gives us the graph of $\calG_i$.
\end{enumerate}
We now prove that the above construction of $U_i^*$ does in fact result in a collection of positions of the vertices in $V^{-(i-1)}$. The associated theorem is stated below with a proof in Appendix \ref{proof_1}.
\setcounter{theorem}{0}
\begin{theorem} \label{thm1}
Given graph $\calG_{i-1}, i \leq m$ in the sequence of graphs transforming a staged tree $\calG_0 = \calS$ to a CEG $\calG_m = \calC$, two situations $v_1, v_2 \in V^{-(i-1)}$ are in the same position if and only if they belong to the same stage and for every $(v_1, v', l)$ there exists a $(v_2, v', l)$ in $\calG_{i-1}$.
\end{theorem}
We now show that the recursion may in fact be stopped for some $0 < r < m$. This optimal stopping point for the recursion is given in Theorem \ref{thm2} with proof in Appendix \ref{proof_2}. 
\begin{theorem}[Optimal stopping] \label{thm2}
In the sequence of graphs transforming a staged tree $\calG_0 = \calS$ to a CEG $\calG_{m} = \calC$ and $m \geq 2$ where $m$ is the depth of $\calS$, the earliest stopping time in this transformation that guarantees the required CEG $\calC$ is the recursion step $r$ such that  $\calG_{r} = \calG_{r-1} \neq \calG_{r-2}$, $0 < r < m$.
\end{theorem}
Theorem \ref{thm3}, with proof in Appendix \ref{proof_3} implies that for every staged tree there is a unique CEG and also that the staged tree can be recovered given this CEG. This is equivalent to saying that no information is lost in transforming a staged tree into a CEG.
\begin{theorem} [Preservation of information] \label{thm3}
The mapping from a staged tree to a CEG is bijective. 
\end{theorem}

\subsection{Related Work}
\label{subsec:related_work}

\cite{silander2013dynamic} presented an algorithm to learn a stratified staged tree and to transform it into an SCEG (although the stratified terminology was not used). Their algorithm is a special case - albeit with no early stopping criterion - of the general algorithm we presented in Section \ref{algorithm}.

They define the structure of a CEG for $n$-dimensional data as a ``layered directed acyclic graph with $n+1$ layers". They assumed that the vertices in layer $k$ correspond to the same variable, say $X_k$. They also assume that from each vertex in layer $k$, there are exactly $r_k$ emanating edges, all entering vertices in layer $k+1$. The stratified staged tree to SCEG transformation algorithm states a weaker form of Theorem \ref{thm1} without a proof and carries out a backward iteration from one layer to the previous one, all the way to the root, by merging situations which satisfy Theorem \ref{thm1}. However, it is easy to see that using their definition of layers, this algorithm fails for non-stratified CEGs where events don't necessarily satisfy a symmetric product space structure. 

We adapt their algorithm so that layer $k$ in their algorithm corresponds to what we defined as set $V^{-k}$ in Section \ref{algorithm}. The main differences between the adapted version of their algorithm and ours is that (1) we provide an optimal stopping criterion which saves on computational effort of searching the entire staged tree, (2) we provide all the necessary proofs for our algorithm. For convenience, call their adapted algorithm the \textit{baseline algorithm} and ours the \textit{optimal time algorithm}.

\section{Experiments} \label{experiments}
Here we compare the performance of the baseline and optimal algorithms on 7 datasets. The first four datasets are from the UCI repository \citep{Dua:2019}. The missing values were removed and sampling zeros were treated as structural. The fifth dataset is from the Christchurch Health and Development Study (CHDS) conducted at the University of Otago, New Zealand (see \cite{fergusson1986social}). The penultimate dataset is from \cite{shenvi2018modelling} and its asymmetric nature can be seen from the CEG in Figure 3 in that paper. The final dataset is an extension of this dataset and has been used in \cite{shenvi2019bayesian}. The last two datasets have structural zeroes and so, they are not stratified and do not have symmetric product space structures. The remaining datasets are stratified. It has also been shown that the last three datasets exhibit context-specific conditional independences \citep{collazo2018chain, shenvi2018modelling, shenvi2019bayesian}.

These experiments were carried out using our Python code \footnote{\url{https://github.com/ashenvi10/Chain-Event-Graphs}} on a 2.9 GHz MacBook Pro with 32GB memory. Our code can handle datasets with structural asymmetries (stored as NaNs or null values) and also provides the capability to manually add sampling zero paths to the tree. It is currently set up to learn the staged tree from the event tree of the dataset using the AHC algorithm. 

\begin{table}[ht]
    \centering
    \begin{tabular}{l|l|l|l|l|l|l}
\multicolumn{1}{c}{\textbf{Dataset}}  & \multicolumn{1}{|c}{\textbf{$\abs{S(\calS)}$}} & \multicolumn{1}{|c}{\textbf{Depth $m$}} & \multicolumn{1}{|c}{\textbf{$T_{\textmd{Baseline}}$}} & \multicolumn{1}{|c}{\textbf{$\abs{V(\calC_{\textmd{Baseline}})}$}} 
& \multicolumn{1}{|c}{\textbf{$T_{\textmd{Optimal}}$}}
& \multicolumn{1}{|c}{\textbf{$\abs{V(\calC_{\textmd{Optimal}})}$}} 
 \\ \hline
       Iris  & 52 & 5 & 1.635 & 42 & 1.414 & 42 \\
     Hayes-Roth & 124 & 5 & 12.118 & 58 & 12.085 & 58  \\
     Balance scale & 327 & 5 & 145.052 & 90 & 143.321 & 90 \\
     Glass & 636 & 10 & 389.272 & 308 & 376.689 & 308 \\
     CHDS & 19 & 4 & 0.586 & 10 & 0.556 & 10 \\
     Falls & 39 & 6 & 1.564 & 27 & 1.453 & 27 \\
     Falls dynamic & 346 & 5 & 585.789 & 242 & 550.990 & 242\\
    \end{tabular}
    \caption{Comparison of the baseline algorithm and the optimal time algorithm.}
    \label{tab:experiment}
\end{table}

Table \ref{tab:experiment} gives for each dataset the number of situations in the staged tree output by the AHC algorithm ($\abs{S(\calS)}$), the maximum depth of the staged tree ($m$) and the time taken (in milliseconds) by the two compacting algorithms ($T_{\textmd{Baseline}}$ and $T_{\textmd{Optimal}}$) as well as the number of positions in the resulting CEG found by the two algorithms ($\abs{V(\calC_{\textmd{Baseline}})}$ and $\abs{V(\calC_{\textmd{Optimal}})}$). From this table we can see that the optimal time algorithm takes less time than the baseline algorithm while arriving at the same CEG as it stops as soon as Theorem \ref{thm2} is satisfied. However, the gain in efficiencies are inversely proportional to the number of symmetries exhibited by the process (see Section \ref{staged_trees}). Thus, if there are more symmetries (more situations in non-trivial stages) across the tree, we need to search across more $V^{-k}$ sets before we arrive at the CEG.

\section{Discussion} \label{discussion}
We have provided a simple iterative backward algorithm along with supporting Python code to transform any staged tree into a CEG. Research in CEGs and their applications has been an increasingly active field in recent years. However, such a general algorithm and proofs of the validity of the staged tree to CEG transformation have been missing in the literature so far. We know through personal correspondence that, a soon to be published, d-separation theorem for CEGs has been developed. Construction of the minimal ancestral CEGs in this theorem follows the same procedure as our algorithm. Hence, automating this process, as we have done, is a very timely development.

\acks{We would like to thank John Horwood and the CHDS research group for the CHDS dataset. We would also like to thank the reviewers whose insightful comments greatly improved the original version. AS was supported by the University of Warwick Chancellor's International Scholarship and the Alan Turing Institute. JQS was supported by the Alan Turing Institute and funded by the EPSRC [grant number EP/K03 9628/1].}

\appendix
\section{Proofs}

\subsection{Proof for Theorem \ref{thm1}} \label{proof_1}
We have a graph $\calG_{i-1}$ belonging to the sequence of graphs converting a staged tree $\calS$ into a CEG $\calC$. This implies that all the vertices in $V^{-j}$, $j = 1,\ldots,i-2$ in $\calG_{i-1}$ represent positions. 

$\Rightarrow$ Given that two situations $v_1, v_2 \in V^{-(i-1)}$ are in the same position. We show that (1) $v_1$ and $v_2$ belong to the same stage; (2) for every $(v_1, v', l)$ there exists a $(v_2, v', l)$ in $\calG_{i-1}$.

If $v_1$ and $v_2$ are in the same position, it is trivially true that they are also in the same stage. Additionally, by the definition of a position, the subtrees rooted at $v_1$ and $v_2$, call them $\calS_{v_1}$ and $\calS_{v_2}$ in the staged tree $\calS$ are isomorphic. Thus also, for every subtree rooted at a child of $v_1$ in $\calS_{v_1}$, there exists an isomorphic subtree rooted at a child of $v_2$ in $\calS_{v_2}$. In fact, stages by definition require that edges with the same estimated conditional transition probability must also have the same edge label. Therefore, there necessarily exists a situation $v_2^{ch}$ along edge $(v_2, v_2^{ch}, l)$ such that the subtree rooted at $v_2^{ch}$ is isomorphic to the subtree rooted at situation $v_1^{ch}$ which is along the edge $(v_1, v_1^{ch}, l)$. Notice that $v_1^{ch}$ and $v_2^{ch}$ belong to the set $V^{-(i-2)}$ in $\calG_{i-2}$. Since their rooted subtrees in $\calS$ are isomorphic, they belong to the same position and are represented by a single vertex, say $v_{1,2}^{ch}$ in $\calG_{i-1}$. The edges $(v_1, v_1^{ch}, l)$ in $\calS_{v_1}$ and $(v_2, v_2^{ch}, l)$ in $\calS_{v_2}$ are represented by edges $(v_1, v_{1,2}^{ch}, l)$ and $(v_2, v_{1,2}^{ch}, l)$ in $\calG_{i-1}$. This result extends to every $(v_1, v', l)$ in $\calG_{i-1}$.

$\Leftarrow$ Given that $v_1, v_2 \in V^{-(i-1)}$ in $\calG_{i-1}$ belong to the same stage and for every $(v_1, v', l)$ there exists a $(v_2, v', l)$ in $\calG_{i-1}$. We need to show that $v_1$ and $v_2$ are in the same position. 

Recall that two situations are in the same position when the subtrees rooted at these vertices in $\calS$ are isomorphic. Since $v_1$ and $v_2$ are in the same stage, they have the same number of emanating edges and also, the edges from $v_1$ and $v_2$ which share the same edge label have the same estimated conditional transition probability. Consider edges $(v_1, v_{1,2}^{ch}, l)$ and $(v_2, v_{1,2}^{ch}, l)$ emanating from situations $v_1$ and $v_2$ in $\calG_{i-1}$ respectively where $v_{1,2}^{ch}$ is the common situation along these two edges. In a tree each vertex has at most one parent. So in the staged tree $\calS$, the position $v_{1,2}^{ch}$ would be represented by two separate vertices, call them $v_1^{ch}$ and $v_2^{ch}$ in the subtrees rooted at $v_1$ and $v_2$ respectively. Thus, the edge $(v_1, v_{1,2}^{ch}, l)$ would be replaced by an edge $(v_1, v_1^{ch}, l)$ in the subtree rooted at $v_1$, call this $\calS_{v_1}$ in $\calS$. Similarly, the edge $(v_2, v_{1,2}^{ch}, l)$ would be replaced by an edge $(v_2, v_2^{ch}, l)$ in $\calS_{v_2}$ which is the subtree rooted at $v_2$ in $\calS$. Since $v_1^{ch}$ and $v_2^{ch}$ are in the same position in $\calG_{i-1}$, they have isomorphic subtrees in $\calS_{v_1}$ and $\calS_{v_2}$. Similarly, the subtrees rooted at the children of $v_1$ and $v_2$ in $\calS_{v_1}$ and $\calS_{v_2}$ respectively are isomorphic whenever the edges from $v_1$ and $v_2$ to their respective children share the same edge label. Since $v_1$ and $v_2$ are in the same stage, the florets $F(v_1)$ in $\calS_{v_1}$ and $F(v_2)$ in $\calS_{v_2}$ are also isomorphic. Thus $\calS_{v_1}$ and $\calS_{v_2}$ are isomorphic and hence, they belong to the same position. 

\subsection{Proof for Theorem \ref{thm2}} \label{proof_2}

Suppose that $0 < r < m$ recursions have taken place and $\calG_{r} = \calG_{r-1} \neq \calG_{r-2}$. We show that $\calG_{r} = \calC$. As the graph of a CEG is the most parsimonious representation of the event tree describing a process, this is equivalent to showing that $\abs{V(\calG_{r})}= \abs{\mathbb{W}} + 1$ where $\mathbb{W}$ is the collection of positions. Graph $\calG_{r-1}$ contains the positions for all situations in $V^{-k}$, $0 \leq k < r-1$. Since $\calG_{r} = \calG_{r-1}$, the problem can be framed as showing that if there are no non-trivial positions in $V^{-(r-1)}$ then there are no non-trivial positions in any of $V^{-k}$, $r \leq k \leq m$. We prove this by contradiction.

Let there be no non-trivial positions in $V^{-{(r-1)}}$. Suppose that two situations $v_1, v_2 \in V^{-r}$ are in the same position and hence, the same stage. This implies that the subtrees of $\calS$ rooted at $v_1$ and $v_2$, say $\calS_{v_1}$ and $\calS_{v_2}$ respectively are isomorphic. Let $v_1^{ch}$ be a child of $v_1$ along the edge $(v_1, v_1^{ch}, l)$ and let $\calS_{v_1^{ch}}$ be the subtree rooted at $v_1^{ch}$. By the definition of a stage, there exists an edge $(v_2, v_2^{ch}, l)$ in $\calS_{v_2}$ with rooted subtree $\calS_{v_2^{ch}}$. The subtrees $\calS_{v_1^{ch}}$ and $\calS_{v_2^{ch}}$ are isomorphic as $\calS_{v_1}$ and $\calS_{v_2}$ are isomorphic. By the definition of a position, $v_1^{ch}$ and $v_2^{ch}$ are in the same position. As $v_1, v_2 \in V^{-r}$, we have that $v_1^{ch}, v_2^{ch} \in V^{-(r-1)}$. This contradicts that there are no non-trivial positions in $V^{-(r-1)}$. A similar argument can be made for any $v_1, v_2 \in V^{-k}$, $r \leq k \leq m$. Since $\calG_{r} = \calG_{r-1}$, $V^{-(r-1)}$ has no non-trivial positions and all the positions in $V^{-k}$, $0 \leq k < r$ have been identified. By the above result, $V^{-k}$, $r \leq k \leq m$ also do not contain any non-trivial positions. Thus $\calG_{r} = \calC$.

We have that $\calG_{r-2} \neq \calG_{r-1} = \calG_r = \ldots = \calG_{m} = \calC$. While stopping at graph $\calG_{r-1}$ gives us the required graph of the CEG, this recursive step is indistinguishable from any of the other $k < r-1$ steps. Hence, the isomorphism of $\calG_{r-1}$ and $\calG_{r}$ is needed to stop the recursions with certainty. Thus the earliest stopping point for the recursion is step $r$ such that $\calG_r = \calG_{r-1} \neq \calG_{r-2}$, $0 < r < m$.

\subsection{Proof for Theorem \ref{thm3}} \label{proof_3}

We prove bijection by proving injection and surjection.

\textbf{Injection:} We prove the injective contrapositive; that is, given staged trees $\calS_1 \neq \calS_2$, we show that their corresponding CEGs $\calC_1$ and $\calC_2$ are not isomorphic. It is straightforward to show that if $\calS_1$ and $\calS_2$ are structurally not isomorphic, then $\calC_1 \neq \calC_2$. Suppose that $\calS_1$ and $\calS_2$ are structurally isomorphic and that they differ only in the colouring of one of their vertices. Let these vertices be $v_1$ with colour $c_1$ in $\calS_1$ and $v_2$ with colour $c_2 \neq c_1$ in $\calS_2$. Since vertices retain their colouring in the CEG, the positions representing $v_1$ and $v_2$ in $\calC_1$ and $\calC_2$ will be coloured by $c_1$ and $c_2$ respectively. Hence, $\calC_1$ and $\calC_2$ will not have colour preserving isomorphism. Additionally, $\calC_1$ and $\calC_2$ will not be structurally isomorphic if either or both of $v_1$ and $v_2$ create non-trivial positions in their respective staged trees as the collection of positions in $\calS_1$ and $\calS_2$ will not be equivalent.

\textbf{Surjection:} From a given CEG $\calC$, construct a staged tree $\calS$ as follows:
\begin{enumerate}[leftmargin=*, topsep = 0pt]
    \itemsep0em
    \item Sort the paths in $\calC_\Lambda$ in ascending order of the length (number of tuples) of the paths.
    \item For each path $\{(c, l)\}$ of length $1$ where $c$ is a colour and $l$ is an edge label, construct an edge from $v_0$, the root of $\calS$ to a new vertex (labelled as $v_i$ where $i$ is an integer index which hasn't been assigned thus far in the construction) and label it $l$. Assign colour $c$ to $v_0$.
    \item In general, for any path of length $k$ given by $\{(c_1, l_1),\ldots,(c_{k-1}, l_{k-1}), (c_k, l_k)\}$, there necessarily exists a path $\{(c_1, l_1),\ldots,(c_{k-1}, l_{k-1})\}$ ending in a vertex, say $v_{k-1}$ in the staged tree constructed so far. To add the $k$th tuple $(c_k, l_k)$ to this path, colour $v_{k-1}$ by $c_k$, add a vertex $v_k$ and construct a directed edge from $v_{k-1}$ to $v_k$ with edge label $l_k$.
\end{enumerate}

This construction results in a tree as it is connected (no vertex - with the exception of the root - is added until it is connected by an edge to an existing vertex) and has no directed cycles (each edge is constructed from an existing vertex to a new vertex). Call this tree $\calT^*$. We prove that $\calT^*$ is the unique staged tree whose transformation, as described in Section \ref{algorithm}, results in our given CEG $\calC$.

Observe that a staged tree in uniquely and unambiguously defined by its underlying event tree and its collection of stages $\mathbb{U}$. The structure of any event tree can be recovered from its set of uncoloured root-to-leaf paths, which is equivalent to the uncoloured root-to-sink paths $\calC_\Lambda$ of the CEG $\calC$. As $\calT^*$ is constructed from the set $\calC_\Lambda$, the uncoloured version of $\calT^*$ is the required underlying event tree for $\calC$. The vertices of $\calT^*$ inherit their colourings from the positions of $\calC$. Recall that colouring of positions in a CEG is indicative of stage memberships. Hence, two positions $w$ and $w'$ with the same colour, say $c$ in $\calC$ are in the same stage. By definition of a stage, $\pmb{\theta}_w = \pmb{\theta}_{w'}$, and for each edge $e = (w, \cdot, l)$ there exists an $e' = (w', \cdot, l)$ such that $\theta(e) = \theta(e')$ in $\calC$. Two vertices $v$ and $v'$ with the colour $c$ in $\calT^*$ either belong to the same position in $\calC$ - without loss of generality assume this is $w$ - or belong to two distinct positions in $\calC$, assume these are $w$ and $w'$. If both $v, v'$ belong to position $w$, then $v$ and $v'$ in $\calT^*$ are created from two separate root-to-$w$ subpaths, say $p$ and $p'$ in $\calC$. Floret $F(v)$ is formed by creating $k$ copies of subpath $p$ and appending each with a distinct $(c, l_i)$ where $i = 1, \ldots, k$ and $l_i$ is the label of the $i$th edge emanating from $w$ in $\calC$. Floret $F(v')$ is constructed in a similar manner. Thus $v$ and $v'$ have the same number of emanating vertices in $\calT^*$ and share the same vertex colour as they satisfy the conditions of being in the same stage by belonging to the same position in $\calC$. This also holds when $v$ and $v'$ belong to positions $w$ and $w'$ respectively, where $w$ and $w'$ share the same colour in $\calC$, with the exception that $p$ will be a root-to-$w$ subpath and $p'$ a root-to-$w'$ subpath. Thus $\calT^*$ is the underlying staged tree of $\calC$ as it has the structure of the event tree of $\calC$ and a collection of stages equivalent to that of $\calC$.

\vskip 0.2in
\bibliography{references}

\begin{thebibliography}{19}
\providecommand{\natexlab}[1]{#1}
\providecommand{\url}[1]{\texttt{#1}}
\expandafter\ifx\csname urlstyle\endcsname\relax
  \providecommand{\doi}[1]{doi: #1}\else
  \providecommand{\doi}{doi: \begingroup \urlstyle{rm}\Url}\fi

\bibitem[Barclay et~al.(2013)Barclay, Hutton, and Smith]{barclay2013refining}
L.~M. Barclay, J.~L. Hutton, and J.~Q. Smith.
\newblock Refining a {B}ayesian network using a chain event graph.
\newblock \emph{International Journal of Approximate Reasoning}, 54\penalty0
  (9):\penalty0 1300--1309, 2013.

\bibitem[Boutilier et~al.(1996)Boutilier, Friedman, Goldszmidt, and
  Koller]{boutilier1996context}
C.~Boutilier, N.~Friedman, M.~Goldszmidt, and D.~Koller.
\newblock Context-specific independence in {B}ayesian networks.
\newblock In \emph{Proceedings of the 12th Conference on Uncertainty in
  Artificial Intelligence}, pages 115--123, 1996.

\bibitem[Carli et~al.(2020)Carli, Leonelli, Riccomagno, and
  Varando]{carli2020r}
F.~Carli, M.~Leonelli, E.~Riccomagno, and G.~Varando.
\newblock The r package stagedtrees for structural learning of stratified
  staged trees.
\newblock \emph{arXiv preprint arXiv:2004.06459}, 2020.

\bibitem[Collazo and Taranti(2017)]{ceg_package}
R.~Collazo and P.~Taranti.
\newblock \emph{ceg: Chain event graph}, 2017.
\newblock URL \url{https://CRAN.R-project.org/package=ceg}.
\newblock R package version 0.1.0.

\bibitem[Collazo et~al.(2018)Collazo, G{\"o}rgen, and Smith]{collazo2018chain}
R.~A. Collazo, C.~G{\"o}rgen, and J.~Q. Smith.
\newblock \emph{Chain event graphs}.
\newblock CRC Press, 2018.

\bibitem[Cowell and Smith(2014)]{cowell2014causal}
R.~G. Cowell and J.~Q. Smith.
\newblock Causal discovery through {MAP} selection of stratified chain event
  graphs.
\newblock \emph{Electronic Journal of Statistics}, 8\penalty0 (1):\penalty0
  965--997, 2014.

\bibitem[Dua and Graff(2019)]{Dua:2019}
D.~Dua and C.~Graff.
\newblock {UCI} machine learning repository, 2019.
\newblock URL \url{http://archive.ics.uci.edu/ml}.

\bibitem[Fergusson et~al.(1986)Fergusson, Horwood, and
  Shannon]{fergusson1986social}
D.~Fergusson, L.~Horwood, and F.~Shannon.
\newblock Social and family factors in childhood hospital admission.
\newblock \emph{Journal of Epidemiology \& Community Health}, 40\penalty0
  (1):\penalty0 50--58, 1986.

\bibitem[Freeman and Smith(2011)]{freeman2011bayesian}
G.~Freeman and J.~Q. Smith.
\newblock Bayesian {MAP} model selection of chain event graphs.
\newblock \emph{Journal of Multivariate Analysis}, 102\penalty0 (7):\penalty0
  1152--1165, 2011.

\bibitem[G{\"o}rgen and Smith(2016)]{gorgen2016differential}
C.~G{\"o}rgen and J.~Q. Smith.
\newblock A differential approach to causality in staged trees.
\newblock In \emph{Conference on Probabilistic Graphical Models}, pages
  207--215, 2016.

\bibitem[G{\"o}rgen and Smith(2018)]{gorgen2018equivalence}
C.~G{\"o}rgen and J.~Q. Smith.
\newblock Equivalence classes of staged trees.
\newblock \emph{Bernoulli}, 24\penalty0 (4A):\penalty0 2676--2692, 2018.

\bibitem[Jabbari et~al.(2018)Jabbari, Visweswaran, and
  Cooper]{jabbari2018instance}
F.~Jabbari, S.~Visweswaran, and G.~F. Cooper.
\newblock Instance-specific {B}ayesian network structure learning.
\newblock \emph{Proceedings of machine learning research}, 72:\penalty0 169,
  2018.

\bibitem[Shenvi and Smith(2019)]{shenvi2019bayesian}
A.~Shenvi and J.~Q. Smith.
\newblock A {B}ayesian dynamic graphical model for recurrent events in public
  health.
\newblock \emph{arXiv preprint arXiv:1811.08872}, 2019.

\bibitem[Shenvi et~al.(2018)Shenvi, Smith, Walton, and
  Eldridge]{shenvi2018modelling}
A.~Shenvi, J.~Q. Smith, R.~Walton, and S.~Eldridge.
\newblock Modelling with non-stratified chain event graphs.
\newblock In \emph{International Conference on Bayesian Statistics in Action},
  pages 155--163, 2018.

\bibitem[Silander and Leong(2013)]{silander2013dynamic}
T.~Silander and T.-Y. Leong.
\newblock A dynamic programming algorithm for learning chain event graphs.
\newblock In \emph{International Conference on Discovery Science}, pages
  201--216. Springer, 2013.

\bibitem[Smith and Anderson(2008)]{smith2008conditional}
J.~Q. Smith and P.~E. Anderson.
\newblock Conditional independence and chain event graphs.
\newblock \emph{Artificial Intelligence}, 172\penalty0 (1):\penalty0 42--68,
  2008.

\bibitem[Su and Zhang(2005)]{su2005representing}
J.~Su and H.~Zhang.
\newblock Representing conditional independence using decision trees.
\newblock In \emph{AAAI}, pages 874--879, 2005.

\bibitem[Zhang and Su(2004)]{zhang2004conditional}
H.~Zhang and J.~Su.
\newblock Conditional independence trees.
\newblock In \emph{European Conference on Machine Learning}, pages 513--524.
  Springer, 2004.

\bibitem[Zhang and Poole(1999)]{zhang1999role}
N.~L. Zhang and D.~Poole.
\newblock On the role of context-specific independence in probabilistic
  inference.
\newblock In \emph{Proceedings of the 16th International Joint Conference on
  Artificial intelligence}, volume~2, pages 1288--1293, 1999.

\end{thebibliography}
\end{document}